\documentclass[pdflatex,sn-mathphys-num]{sn-jnl}

\usepackage{graphicx}%
\usepackage{multirow}%
\usepackage{amsmath,amssymb,amsfonts}%
\usepackage{amsthm}%
\usepackage{mathrsfs}%
\usepackage[title]{appendix}%
\usepackage{xcolor}%
\usepackage{textcomp}%
\usepackage{manyfoot}%
\usepackage{booktabs}\let\cline\cmidrule%
\usepackage{algorithm}%
\usepackage{algorithmicx}%
\usepackage{algpseudocode}%
\usepackage{listings}%
\usepackage{makecell}%
\usepackage{float}%
\usepackage{dcolumn}%
\usepackage{lipsum}

\PassOptionsToPackage{hyphens}{url}\usepackage{hyperref}%

\begin{document}

\title[Energy Efficient Text Classification]{Comparing energy consumption and accuracy in text classification inference}


\author{\fnm{Johannes} \sur{Zschache}}\email{johannes.zschache@uba.de}

\author*{\fnm{Tilman} \sur{Hartwig}}\email{tilman.hartwig@uba.de}

\affil{\orgdiv{Application Lab for AI and Big Data}, \orgname{German Environment Agency}, \orgaddress{\street{Alte Messe 6}, \city{Leipzig}, \postcode{04103}, \state{Saxony}, \country{Germany}}}



\abstract{The increasing deployment of large language models (LLMs) in natural language processing (NLP) tasks raises concerns about energy efficiency and sustainability. While prior research has largely focused on energy consumption during model training, the inference phase has received comparatively less attention. This study systematically evaluates the trade-offs between model accuracy and energy consumption in text classification inference across various model architectures and hardware configurations. Our empirical analysis shows that in some contexts the best-performing model in terms of accuracy can also be energy-efficient. While LLMs tend to consume significantly more energy than traditional machine learning models, they show the same or even lower levels of accuracy in our zero-shot classification setting. We observe substantial variability in inference energy consumption ($<$mWh to $>$kWh), influenced by model type, model size, and hardware specifications. Additionally, we find a strong correlation between inference energy consumption and model runtime, indicating that execution time can serve as a practical proxy for energy usage in settings where direct measurement is not feasible. Our findings demonstrate that energy efficiency and accuracy represent distinct evaluation dimensions that do not necessarily align. We argue that sustainable AI development requires systematic evaluation of both performance and resource efficiency.}




\maketitle

\begin{figure}[H]
\centering
\makebox[\linewidth][c]{%
  \includegraphics[height=\textheight]{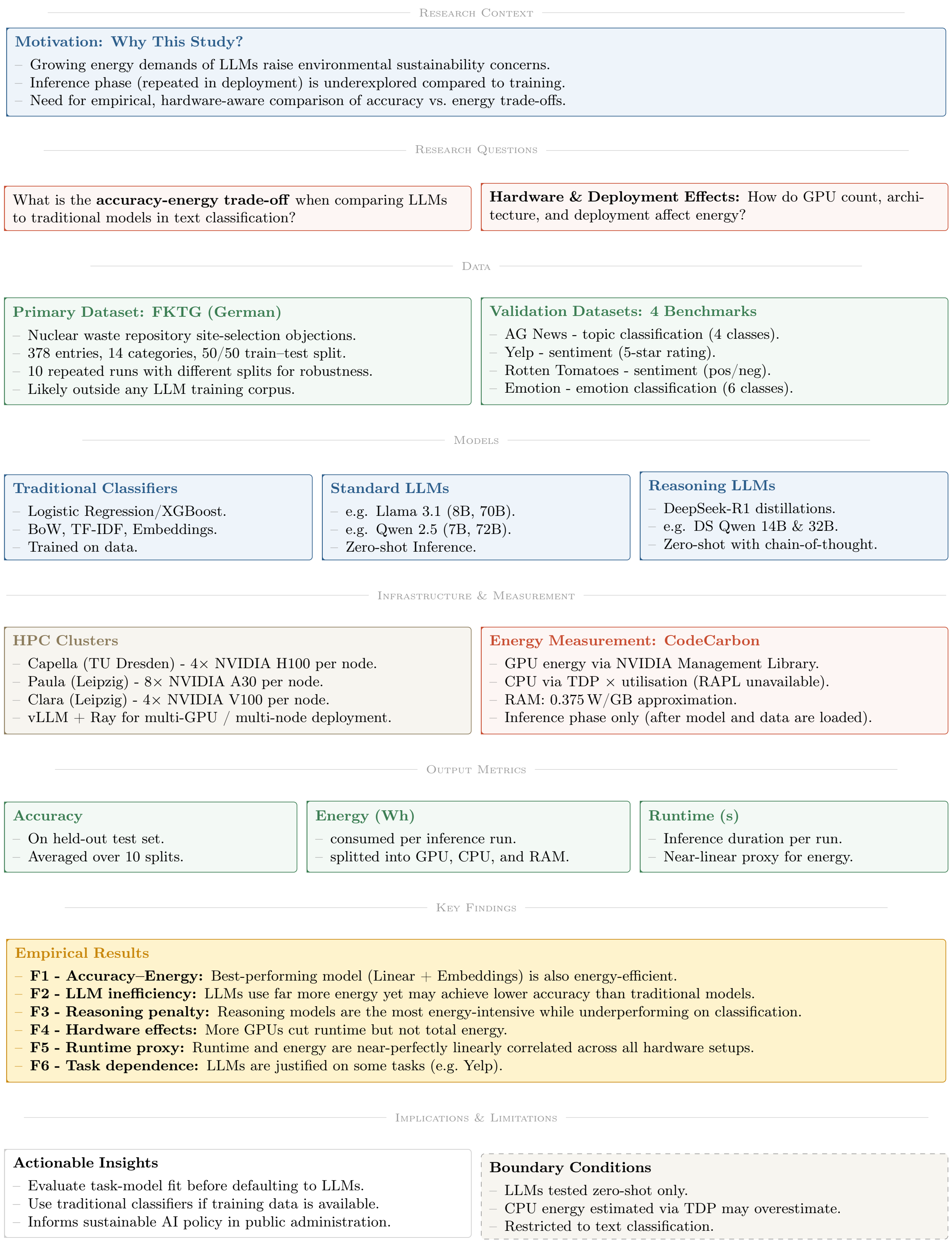}%
}
\caption{Logical Framework Diagram.}
\label{fig:LFD}
\end{figure}

\section{Introduction}\label{sec1}

Artificial intelligence (AI) systems, particularly large language models (LLMs), have driven remarkable progress in Natural Language Processing (NLP) applications. This development has been enabled by the transformer architecture \citep{vaswani2017} and exemplified by the emergence of large-scale models such as GPT-3 \citep{brown2020}, which have significantly advanced task performance. However, this progress has come at a cost: the escalating energy demands of AI systems pose significant environmental and computational challenges. Data centers that support AI computations are major electricity consumers, often dependent on fossil fuels, thereby contributing to greenhouse gas emissions \citep{lacoste2019,axenbeck2025}. This increasing energy demand challenges global climate objectives such as the Paris Agreement \citep{UN2015Paris} and the United Nations' Sustainable Development Goals (SDGs), specifically Goal 13 on climate action \citep{UN2015SDGs}. Consequently, designing energy-efficient AI systems is imperative for aligning technological advancements with sustainability goals. Moreover, beyond sustainability, energy-efficient models offer additional advantages, including reduced hardware requirements, lower financial costs, and faster inference times.

When evaluating machine learning models, most studies concentrate on the quality of the model responses by tracking e.g. the accuracy, the RMSE, or other measures. And even if the energy consumption is taken into account, prior research has mainly focused on the training phase \citep{strubell2019,patterson2021,luccioni2023}. The inference phase, which is repeatedly executed in real-world deployments, has received comparatively less attention. However, energy efficiency during the operational phase is an increasingly relevant topic as LLM applications become ubiquitous and LLMs are trained to use additional test-time compute to improve performance \citep{openai2024,deepseekai2025deepseekr1}. Addressing this gap, we present a systematic study on the energy consumption of language models during inference, providing actionable insights for balancing accuracy with efficiency.

A particularly popular machine learning task is text categorization, a task that lightweight models have been shown to handle effectively. For instance, \citet{Joulin2017} show that a simple classifier built on word embeddings is often as accurate as deep learning classifiers. Despite this, some authors argue for the use of pre-trained LLMs for text classification because it reduces the need for model training and simplifies data preprocessing \citep{wang2024}. Additionally, popular software tutorials promote LLMs for classification tasks \citep{langchain_classification,lamini_cat}, further encouraging their use even when more efficient alternatives exist. In order to justify the usage of LLM in relatively simple tasks such as text categorization, we advocate a consequent comparison of a model's response quality to its energy efficiency.

Using a practical use case from public administration, our study empirically analyzes trade-offs between model accuracy and energy consumption across various language models and hardware configurations. We find that high accuracy does not necessarily require high energy expenditure: the best-performing models can also be energy-efficient. In our zero-shot classification setup, LLMs generally exhibit substantially higher energy consumption than traditional models without corresponding gains in accuracy. Generally, we observe significant variability in inference energy consumption, influenced by model type, model size, and hardware specifications. Additionally, energy consumption during inference correlates strongly with model runtime, suggesting that execution time can serve as a valuable proxy for energy usage when direct measurement is unavailable. The logical workflow of our study is summarized in Fig.~\ref{fig:LFD}.

Our findings have implications for researchers, practitioners, and policymakers advancing sustainable AI development \citep{kaack2022, luccioni2025}. We argue that model assessment should incorporate systematic evaluation of inference efficiency alongside traditional performance metrics. By presenting a reproducible framework for such evaluation, we contribute to the ongoing discourse on AI's environmental impact and demonstrate how energy considerations can be integrated into model selection and deployment decisions for NLP applications.  To support reproducibility and enable further research, we have published the code, measurement data, and analysis online\footnote{\url{https://gitlab.opencode.de/uba-ki-lab/llm-testframework}}.

\section{Previous research}\label{sec2}

Research on the environmental impact of machine learning (ML) has primarily focused on the energy consumption and carbon emissions produced during the training phase of large-scale models. Most famously, \citet{strubell2019} quantify the carbon footprint of NLP models, revealing that the training of a single large-scale transformer model can emit as much carbon as five cars over their entire lifetimes (their measurements include thousands of hyperparameter tuning jobs, which makes it difficult to disentangle model-inherent efficiency from experimental setup). This seminal work spurred further investigations into the environmental costs of training neural networks, including large language models \citep{patterson2021,patterson2022,luccioni2023,falk2025carbonc}.

While training remains a significant contributor to energy consumption, recent studies have begun to focus on the inference phase. \citet{samsi2023} highlighted the substantial energy demands of LLM inference but did not explore the relationship between energy consumption and task-specific performance. \citet{liu2022} underscore the importance of evaluating NLP models not just on efficiency metrics but also on accuracy by introducing the Efficient Language Understanding Evaluation (ELUE) benchmark. ELUE aims to establish a Pareto frontier that balances performance and efficiency. It includes various language understanding tasks, facilitating fair and comprehensive comparisons among models. However, the framework adopts number of parameters and FLOPs as the metrics for model efficiency, disregarding hardware specific factors. Similarly, \citet{chien2023} estimate the energy consumption associated with the inference phase of generative AI applications based on the output word count and several assumptions about the application such as the number of FLOPS per inference and the sampling rate.
More recently, \citet{oviedo2025energyuseaiinference} develops a bottom-up, production-oriented token-throughput methodology to estimate per-query energy and shows that production optimizations can substantially lower Wh-per-query consumption. 

In contrast, we promote energy-efficient NLP models by the direct measurement of the power consumed during inference. Hence, our work follows the approach of the SustaiNLP 2020 shared task \citep{Wang2020}. SustaiNLP demonstrated that substantial energy savings are achievable with minimal performance loss. While this study was limited to the performance of a couple of small language models on a single benchmark, we extend these efforts to a greater number of partially very large models deployed to a practical inference scenario. 

This makes our study very similar to the one by \citet{alizadeh2025}, who investigated the trade-offs between accuracy and energy consumption when deploying large language models (LLMs) for software development tasks. Besides the finding that larger LLMs with higher energy consumption do not always yield significantly better accuracy, the authors demonstrated that architectural factors, such as feedforward layer size and transformer block count, directly correlate with energy usage.

Finally, \citet{LuccioniEtAl2024} provide one of the most comprehensive analyses of energy consumption during ML model inference. Their study systematically compared the energy costs of 88 models across 10 tasks and 30 datasets, including both smaller task-specific and larger multi-purpose models. They found that the larger models are orders of magnitude more energy-intensive than smaller task-specific ones, especially for tasks involving text and image generation. Furthermore, their research underscores the variability in energy consumption across tasks and model architectures. The authors advocate for increased transparency and sustainable deployment practices, emphasizing that the environmental costs of deploying large, multi-purpose AI systems must be carefully weighed against their utility.

In summary, recent work has begun to examine inference energy more systematically, including operational-scale analyses \citep{gupta2020chasingcarbon,Elsworth25}, model-specific footprint estimations \citep{luccioni2022}, and emerging benchmarks such as ML.ENERGY \citep{mlenergy2025} and the AI Energy Score \citep{huggingfaceAIEnergyScore}. These efforts highlight the importance of measuring inference-time resource usage but differ from our study in their focus on macro-level infrastructure, generative workloads, or automated benchmarking pipelines rather than comparative, task-specific measurement across heterogeneous models and hardware setups.

\section{Data and methods}\label{sec3}

Our experiments are inspired by an occasionally occurring use case in public administration: the management of objections that are submitted by the population. Due to a potentially very large amount of submissions, an automatic preprocessing of the objections is of high value. One of the possible steps of an automated workflow is to categorize each submission for optimal forwarding to the responsible department.

The data of our study originates from the process of selecting a repository site for high-level radioactive waste in Germany. During the first phase, sub-areas were identified and discussed in a process called FKTG (Fachkonferenz Teilgebiete). The statements from the population were categorized, processed and published as the FKTG-dataset\footnote{\url{https://beteiligung.bge.de/index.php}}. The text of the submission is given by the column `Beitrag' (input). The column `Themenkomplex' (topic) contains the category of the text. Before publication, the texts were manually cleaned by e.g.\ removing any personal information or templated repetitions.

We scraped the dataset from the website and restricted it to entries for which the topic occurs at least 10 times, resulting in 378 entries across 14 categories. While this dataset size is relatively small compared to typical benchmarks in NLP research, it is representative of real-world scenarios in public administration where labeled data is often limited and expensive to obtain. The domain-specific nature of the dataset - focusing on technical and administrative language related to radioactive waste management - presents both a limitation and an opportunity: while our findings may not generalize to all text classification tasks, they provide valuable insights for specialized domains where energy-efficient solutions are particularly important due to resource constraints. Additionally, this specialized vocabulary and context allows us to evaluate how different model types handle domain-specific language without extensive training data, a common challenge in governmental and scientific applications.
To address dataset-specific limitations and to test whether our findings persist under more challenging conditions, we complement the FKTG experiments with four additional benchmark datasets representing substantially different difficulty profiles (see Sec.~\ref{sec:other-datasets}).

The dataset was split into half: 189 entries for training and 189 entries for testing. This unconventional 50:50 split, rather than the standard 70:30 or 80:20 split, was deliberately chosen for several reasons. First, with only 378 total entries across 14 categories (averaging 27 instances per category), a larger test set ensures statistical reliability of our performance metrics, particularly for minority classes. Second, since our primary focus is on inference energy consumption rather than achieving maximum accuracy through extensive training, we prioritize having a representative test set that can yield robust energy measurements across all categories. Third, this split creates a more challenging scenario for traditional models, allowing us to assess whether their potential accuracy disadvantage compared to zero-shot LLMs is offset by their energy efficiency benefits - a key trade-off in our analysis. To mitigate concerns about reduced training data, each experiment was repeated 10 times with different random train-test-splits, ensuring that our results reflect average performance rather than artifacts of a particular split. To increase comparability, every experiment was run with the same 10 train-test-splits.

An experiment run consists of a training phase and a testing phase. Since large language models have been argued to be applicable to text categorization without training (zero-shot), we omit the training phase for these models and apply LLMs without fine-tuning. We report the energy consumption and accuracy only for the test phase as averages over all 10 runs.

\subsection{Traditional models}

Besides LLMs, we initially run the experiments with lightweight NLP models that we call traditional because they have been used for categorization tasks long before LLMs existed. Specifically, we use a linear model (logistic regression) and a gradient boosting algorithm (xgboost). Logistic regression is a simple, interpretable model that estimates the probability of a class based on a linear combination of input features. XGBoost (Extreme Gradient Boosting) is an efficient, scalable machine-learning algorithm that combines predictions from multiple decision trees to improve accuracy.

We consider three different types of features: bag-of-words (BoW), term frequency-inverse document frequency (TF-IDF), and a pretrained multilingual sentence embedding. For text preprocessing and feature extraction, we rely on the default implementations provided by scikit-learn and sentence-transformers libraries to ensure reproducibility and represent common practice in applied NLP scenarios. For BoW features, we use `sklearn.feature\_extraction.text.CountVectorizer' with default parameters, which automatically handles the following preprocessing steps: (1) conversion to lowercase, (2) tokenization using word boundaries while preserving tokens of at least 2 characters, and (3) removal of punctuation at word boundaries. No stop word removal or stemming is applied by default, allowing the model to learn from the full vocabulary including function words that may be informative in our domain-specific context. The resulting BoW representation creates sparse vectors counting word occurrences without considering word order.

For TF-IDF features, we employ `sklearn.feature\_extraction.text.TfidfVectorizer' with character-level n-grams rather than word-level features. Specifically, we configure the vectorizer to extract 2-gram and 3-gram character sequences (analyzer='char', ngram\_range=(2,3)), which effectively captures morphological patterns and handles German compound words common in technical administrative texts. The TF-IDF weighting scheme automatically adjusts feature importance based on document frequency, down-weighting common character sequences while emphasizing distinctive patterns. The default parameters include: minimum document frequency of 1, no maximum document frequency limit, and L2 normalization of the resulting feature vectors. This character-level approach is particularly suitable for German text, where compound words and domain-specific terminology might not be well-represented in standard tokenization approaches.

For the multilingual sentence embedding approach, we use the SentenceTransformer class from the sentence-transformers library with the pretrained model `paraphrase-multilingual-mpnet-base-v2'\footnote{\url{https://huggingface.co/sentence-transformers/paraphrase-multilingual-mpnet-base-v2}}. This model accepts raw text input without requiring explicit preprocessing, as all necessary text normalization is handled internally by the model's tokenizer. The model automatically performs subword tokenization using SentencePiece, handles texts up to 512 tokens (though none of our documents exceeded this limit), and generates dense 768-dimensional vectors that capture semantic meaning across languages. We apply this embedding without fine-tuning on our training data, using the .encode() method directly on the raw text inputs. Importantly, sentence embeddings are generated dynamically for each sample during inference rather than being precomputed, ensuring consistent experimental conditions with the LLM evaluation pipeline and allowing accurate measurement of the complete computational cost including embedding generation.

Both traditional classification models (logistic regression and XGBoost) are trained using the default parameters provided by sklearn.linear\_model.LogisticRegression and xgboost.XGBClassifier, ensuring a fair comparison focused on the impact of feature representations rather than hyperparameter optimization.

\subsection{Large language models}
Large language models (LLMs) were applied without training (zero-shot) using the test set only. This zero-shot approach was chosen for several reasons: (1) it reflects real-world deployment scenarios where fine-tuning may be impractical due to computational costs or data scarcity, (2) it allows us to measure the true inference energy cost without conflating it with fine-tuning overhead, and (3) it provides a fair comparison point for understanding the energy-accuracy trade-off when models are used out-of-the-box. However, we acknowledge several limitations of this methodological choice. Zero-shot prompting may not fully exploit the LLMs' capabilities, as these models could potentially achieve higher accuracy through few-shot learning (providing examples in the prompt) or parameter-efficient fine-tuning methods like LoRA.

Recent empirical work has shown that the choice between zero‑shot and few‑shot prompting yields a non-trivial trade-off between predictive performance and inference cost. For instance, in a systematic evaluation across sentiment analysis, NLI, and other classification tasks using multiple LLMs, \citet{Atetedaye2025_FewShotVsZeroShot} reports that few‑shot prompting generally outperforms zero‑shot, especially for classification and structured-output tasks, while incurring substantially increased inference overhead in terms of token cost, latency, and compute usage, however ``marginal accuracy improvements may not justify doubling inference time". A recent survey of NLP zero‑ and few‑shot approaches \citep{Ramesh2025_ZeroShotFewShotSurvey} finds that zero‑shot methods remain a viable, computationally efficient choice for text classification, particularly under resource constraints or when labeled data is scarce.

While few-shot prompting or fine-tuning could potentially improve accuracy, we opted for zero-shot prompting to prioritize computational efficiency and reproducibility. A rigorous comparison would require extensive experimentation across multiple configurations (e.g., varying numbers of examples, selection strategies, and training regimes), each affecting both performance and cost in task-dependent ways. Moreover, full fine-tuning was deemed incompatible with our focus on inference energy, as it would require reporting training energy costs that vary dramatically based on hardware and optimization strategies. Instruction tuning and retrieval-augmented generation (RAG) approaches, while promising, would introduce additional system components that would obscure the fundamental energy characteristics of the base models themselves.

Our prompt design (Appendix \ref{secApp:LLMPrompt}) adds approximately 100-200 tokens per input for task instructions and category specifications - a 2-3x increase over raw text length. While this overhead contributes to LLMs' higher energy consumption, it is (1) necessary for zero-shot classification to function properly, (2) consistent across all LLM experiments ensuring fair comparison, and (3) representative of real-world deployment where task instructions are unavoidable. Attempts to use more concise prompts resulted in degraded accuracy and format compliance issues, suggesting this token overhead represents a fundamental trade-off in zero-shot LLM deployment rather than an optimization opportunity. Nevertheless, the performance of zero-shot models is highly sensitive to prompt engineering; while we used a standardized prompt template, alternative prompt formulations might yield different results. 

To ensure fair comparison across models, we constrained the LLM outputs to generate only a single category label from the predefined set of 14 categories. As detailed in Appendix \ref{secApp:LLMPrompt}, the prompt explicitly instructs models to respond with only the category name without additional explanation or reasoning. This constraint resulted in highly consistent output lengths across all experimental runs: for standard models, the generated responses varied by at most one or two tokens, corresponding to the minor differences in category name lengths (e.g., ``Endlagerkonzept" versus ``Verfahren"). An important exception is the Deepseek distillation models, which employ a reasoning mechanism that generates additional internal ``thinking" tokens before producing the final answer. These thinking tokens, while not visible in the final output, contribute substantially to the total token count and thus energy consumption - in some cases generating 50-200 additional tokens during the reasoning process. We include these thinking tokens in our energy measurements as they represent an integral part of the models' test-time compute strategy, providing insight into the energy cost of enhanced reasoning capabilities. This variation in internal processing highlights an important dimension of the energy-accuracy trade-off. 

Table \ref{tbl-llm-selection} gives the names and sources of the models used. We use the term ``Large Language Model" (LLM) to refer to transformer-based language models with at least 1 billion parameters that are primarily trained for text generation using a decoder-only architecture. Under this definition, the models listed in Table~\ref{tbl-llm-selection} all qualify as LLMs. In contrast, we refer to smaller encoder-only models like the sentence-transformers model \texttt{paraphrase-multilingual-mpnet-base-v2} (with 278M parameters) as ``embedding models" rather than LLMs, to maintain this distinction.

The LLMs were selected by the following criteria:
\begin{itemize}
\item availability on Huggingface (ensuring reproducibility and standardized loading procedures)
\item support of German language (verified through model documentation and preliminary testing)
\item capability of processing the \texttt{dspy}-prompt without exceeding context limits or formatting errors (see Appendix \ref{secApp:LLMPrompt})
\item diversity in model sizes (from under 1B to over 70B parameters) to examine the relationship between model scale and energy consumption
\end{itemize}

Additionally, Jamba Mini 1.5 was chosen as a model with an alternative architecture that includes next to transformer also Mamba layers (a state-space model), allowing us to assess whether architectural innovations impact energy efficiency. The DeepSeek distillations (DS) were added to include models with reasoning capabilities through test-time compute, representing recent advances in inference-time optimization. This selection strategy ensures coverage of different model families, sizes, and architectures while maintaining practical constraints for energy measurement in controlled conditions.

\begin{table}[h]
\centering
\begin{tabular}{ll}
\hline
Model & Link \\ 
\hline
Llama 3.1 8B& \url{https://huggingface.co/meta-llama/Meta-Llama-3.1-8B-Instruct} \\ 
Llama 3.1 70B & \url{https://huggingface.co/meta-llama/Meta-Llama-3.1-70B-Instruct} \\ 
Qwen 2.5 7B & \url{https://huggingface.co/Qwen/Qwen2-7B-Instruct} \\ 
Qwen 2.5 72B & \url{https://huggingface.co/Qwen/Qwen2-72B-Instruct} \\ 
Phi 3.5 Mini & \url{https://huggingface.co/microsoft/Phi-3.5-mini-instruct} \\ 
Phi 3.5 MoE & \url{https://huggingface.co/microsoft/Phi-3.5-MoE-instruct} \\ 
Jamba Mini 1.5 & \url{https://huggingface.co/ai21labs/AI21-Jamba-1.5-Mini} \\ 
DS Qwen 14B & \url{https://huggingface.co/deepseek-ai/DeepSeek-R1-Distill-Qwen-14B} \\ 
DS Qwen 32B & \url{https://huggingface.co/deepseek-ai/DeepSeek-R1-Distill-Qwen-32B} \\
DS Llama 8B & \url{https://huggingface.co/deepseek-ai/DeepSeek-R1-Distill-Llama-8B} \\ 
DS Llama 70B & \url{https://huggingface.co/deepseek-ai/DeepSeek-R1-Distill-Llama-70B} \\ 
\hline
\end{tabular}
\caption{Selection of large language models}\label{tbl-llm-selection}
\end{table}

\subsection{Computing Resources}

We used different computing systems for a comparative analysis of energy efficiency across diverse hardware architectures. This enables the assessment of how architectural differences - especially GPU tensor core capabilities - affect the inference speed and power usage. A diversity in computational infrastructure is crucial for generalizing findings across different environments and ensuring the validity and replicability of experimental results in machine learning research. Furthermore, insights gained from using multiple platforms contribute to optimizing resource allocation strategies and improving cost-effectiveness in large-scale machine learning projects.

To run our experiments, we were granted access to the high-performance computing (HPC) systems of TUD Dresden University of Technology\footnote{\url{https://doc.zih.tu-dresden.de/}} and Leipzig University\footnote{\url{https://www.sc.uni-leipzig.de/}}. For GPU-accelerated computing, three different systems are available named \texttt{Capella}, \texttt{Paula}, and \texttt{Clara} (see Table \ref{tab:hpc}). The main difference for our study is the GPU: while a node on the \texttt{Capella} cluster is equipped with 4 x H100, there are 8 x A30 on each node on \texttt{Paula} and 4 x V100 on \texttt{Clara}. This means that a large model such as Llama 3.1 70B or Qwen 2.5 72B fits on a single node of \texttt{Capella} (requiring 2 GPUs) or \texttt{Paula} (requiring all 8 GPUs) but takes up two nodes of the \texttt{Clara} cluster (assuming a 16-bit floating point representation of the parameters).

\begin{table}[h]
    \centering
    \begin{tabular}{l|ccc}
      \hline
      Cluster & \thead{Capella} & \thead{Paula} & \thead{Clara} \\
      \hline
      \hline
      HPC center & \makecell{TUD Dresden\\University of Technology} & Leipzig University & Leipzig University \\
      \hline
      number of nodes & 144 & 12 & 6 \\
      \hline
      CPU per node &  \makecell{2 x AMD (32 cores)\\2.7GHz} &  \makecell{2 x AMD (64 cores)\\ 2.0GHz} & \makecell{1 x AMD (32 cores)\\ 2.0GHz}\\
      \hline
      RAM per node &  768 GB & 1 TB & 512 GB \\
      \hline
      GPU per node & \makecell{4 x NVIDIA H100\\(94 GB)} & \makecell{8 x NVIDIA A30\\(24 GB)} & \makecell{4 x NVIDIA V100\\(32 GB)} \\
      \hline
      \makecell{single GPU max\\power consumption} & 700W & 165W & 250W \\
      \hline
    \end{tabular}
    \caption{HPC Resources}
    \label{tab:hpc}
\end{table}

LLMs were deployed using the \texttt{vllm} library\footnote{\url{https://github.com/vllm-project/vllm}}, which runs on a ray cluster\footnote{\url{https://www.ray.io/}} for multi-node computations. If a model is too large to be deployed on a single GPU, the model weights are distributed over multiple GPUs, which allow for a parallel computation of the activations \citep[c.f. tensor model parallelism (TMP) in][pp.16]{bai2024}. In cases where two computing nodes are needed, the model is split into two parts and executed sequentially \citep[c.f. pipeline model parallelism (PMP) in][p.17]{bai2024}: first the model part on the first node and then the model part on the second node.

To ensure consistent and realistic energy measurements, we employed a warm-start inference scenario for all experiments. Specifically, for LLMs, we initialized the vLLM server once before beginning measurements, allowing model weights to be loaded into memory and any necessary compilation or optimization steps to complete. All 189 test samples within each experimental run were then processed using the same server instance without restart, reflecting typical production deployments where models remain loaded for batch processing. For traditional models, the trained model was similarly loaded once into memory before processing the entire test set. This warm-start approach excludes one-time initialization costs (such as model loading from disk, memory allocation, and initial GPU kernel compilation) from our measurements, focusing instead on the steady-state inference energy that would dominate in real-world applications where models serve multiple requests. We acknowledge that cold-start scenarios—where models are loaded fresh for each prediction—would incur additional energy costs, particularly for large LLMs that require substantial memory transfers. However, such deployment patterns are uncommon in practice due to their inefficiency, and including cold-start overhead would obscure the fundamental differences in inference efficiency between model architectures, which is the primary focus of our analysis.

The energy consumption and the runtime of the inference phase were measured using the CodeCarbon package\footnote{\url{https://github.com/mlco2/codecarbon}}. This package uses the NVIDIA Management Library (NVML) and the Intel RAPL files to track the power usage of the GPU and CPU\footnote{\url{https://mlco2.github.io/codecarbon/methodology.html\#power-usage}}. The power consumption of the memory is flatly added with 0.375W/GB of memory used. In settings where the model is deployed on more than one node, the inference duration is taken as the maximum, and the energy is taken as the sum over all nodes.

Due to the unavailability of Running Average Power Limit (RAPL) interface data on our computing infrastructure, CPU energy consumption was estimated using Thermal Design Power (TDP) specifications combined with CPU utilization metrics\footnote{\url{https://mlco2.github.io/codecarbon/methodology.html\#cpu-metrics-priority}}. We acknowledge a key limitation of this approach: TDP reflects maximum sustained power draw, and on shared or modern CPUs with dynamic voltage/frequency scaling, actual consumption may be substantially lower. Since traditional ML models in our study are predominantly CPU-bound, this overestimation applies asymmetrically — it inflates the reported energy consumption of traditional models while leaving GPU-bound LLM measurements largely unaffected. Critically, this means our methodology works against our primary finding: any overestimation of CPU energy makes traditional models appear less efficient relative to LLMs than they may actually be. Therefore, the efficiency advantages of traditional models reported here should be interpreted as conservative lower bounds. Access to RAPL or direct power measurement would likely reduce the absolute energy values for CPU-bound models and potentially reveal even larger efficiency differences than those reported.

Various software tools have been created to monitor energy consumption during the application of machine learning models\footnote{\url{https://github.com/tiingweii-shii/Awesome-Resource-Efficient-LLM-Papers?tab=readme-ov-file\#\%EF\%B8\%8F-energy-metrics}}. Similarly to CodeCarbon, Carbontracker \citep{anthony2020carbontracker} and experiment-impact-tracker \citep{Henderson2020} estimate energy consumption by monitoring hardware usage. In some settings, CodeCarbon is considered more accurate, delivering values closer to those obtained using physical wattmeters \citep{Bouza2023}. Although comparing different tools of energy monitoring is beyond the scope of our paper, one clear disadvantage of CodeCarbon is that the CPU energy cannot be measured for individual cores. This leads to an overestimation of the CPU energy in shared systems such as our infrastructure. While this is negligible in GPU-heavy computations such as LLM inference, it should be kept in mind when interpreting the following results.

\section{Results}\label{sec5}

For each model, we report the accuracy, energy consumption, and duration of the inference. The energy consumption and duration were measured only for the inference step, i.e., after the model and data were loaded. An inference run involves classifying 189 text samples from a test set. All tables and figures present the average results over 10 runs on different test sets, with the same 10 test sets used for each model. The measurement variance was generally low: $< 0.002$ for accuracy, and $< 0.2$\,dex for both energy consumption and duration (logarithmically scaled to base 10).

\begin{figure}[h]
\centering
\includegraphics[width=\textwidth]{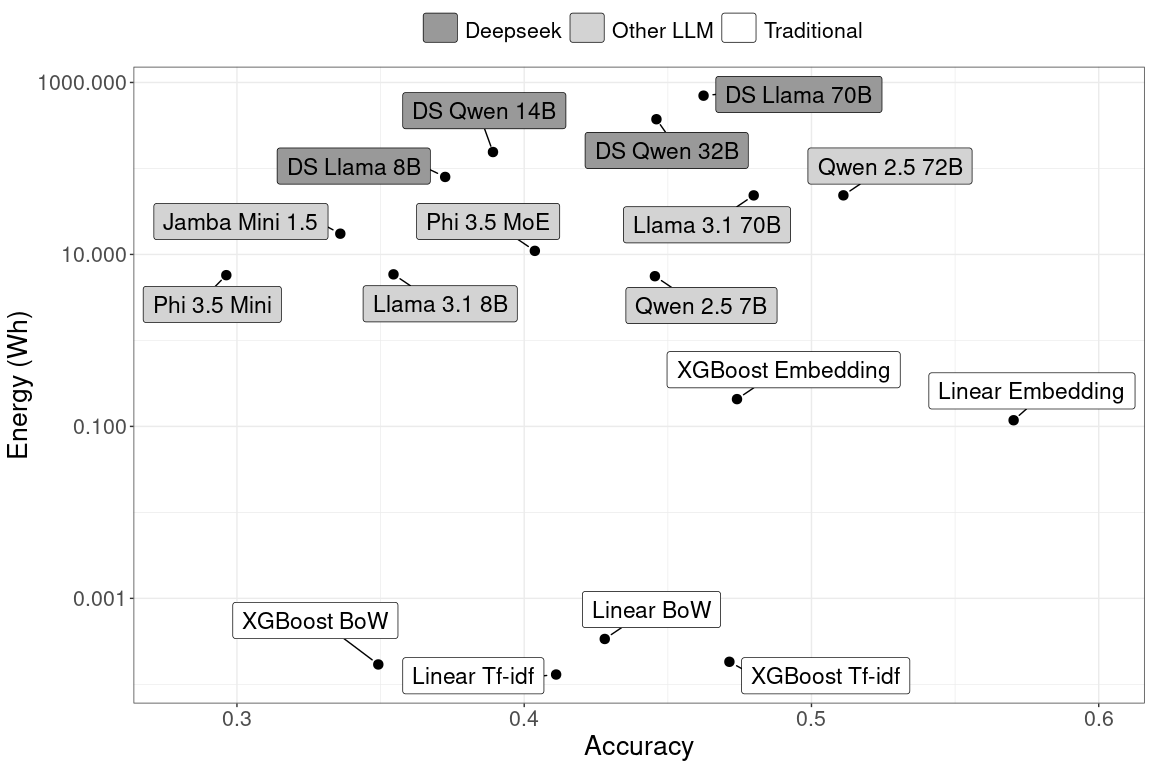}
\caption{Accuracy-energy-trade-off of all models for the inference task on the \texttt{Capella} system (single node). The energy consumption for the same task spans over six orders of magnitude with traditional models are always more energy‑efficient than any of the evaluated LLMs and reasoning models are most energy-consuming. The best model for this specific task is a traditional model (Linear Embedding) with moderate energy consumption.}
\label{fig:inference-all}
\end{figure}

Figure \ref{fig:inference-all} illustrates the trade-off between energy consumption and accuracy across all models. For these experiments, a single node of the \texttt{Capella} system was used. The minimum number of H100 GPUs required varies by model (see Table \ref{tbl:inference-all}).

The highest accuracy was achieved by a traditional linear model using pre-trained sentence embeddings. Notably, even the most energy-efficient model - a linear model with TF-IDF features - outperformed several large language models. Among LLMs with relatively high accuracy, the best small model (Qwen 2.5 7B) consumes seven times less energy than the most accurate model (Qwen 2.5 72B), with only a minor accuracy reduction of 0.07 points. Deepseek models, despite their extensive reasoning processes during inference, exhibit lower accuracy than non-reasoning LLMs while consuming significantly more energy and taking longer to complete inference. 

We acknowledge that our comparison evaluates LLMs in a zero-shot setting while traditional models are fine-tuned on task data, creating an asymmetry in experimental conditions. Our methodological choice represents a conscious trade-off, prioritizing computational efficiency, scalability, and reproducibility, while accepting that fine-tuned LLMs might narrow the performance gap observed in our zero-shot evaluation.

The visually illustrated differences in accuracy and energy consumption were more deeply analysed by a pairwise t-test with corrections for multiple testing (Holm method). In regard to the accuracy, the linear model using pre-trained sentence embeddings is significantly better than all other models. When looking at the energy consumption, it is significantly more efficient than all LLMs (the test results are available at the paper's repository\footnote{\url{https://gitlab.opencode.de/uba-ki-lab/llm-testframework/-/blob/main/analysis/energy_llm.md}}). 

\subsection{Analysis of hardware settings}

This section analyzes the impact of different hardware configurations (see Tab.~\ref{tab:hpc}) on energy consumption. We focus on GPU usage due to its dominant role in machine learning inference. 

\begin{figure}[h]
\centering
\includegraphics[width=\textwidth]{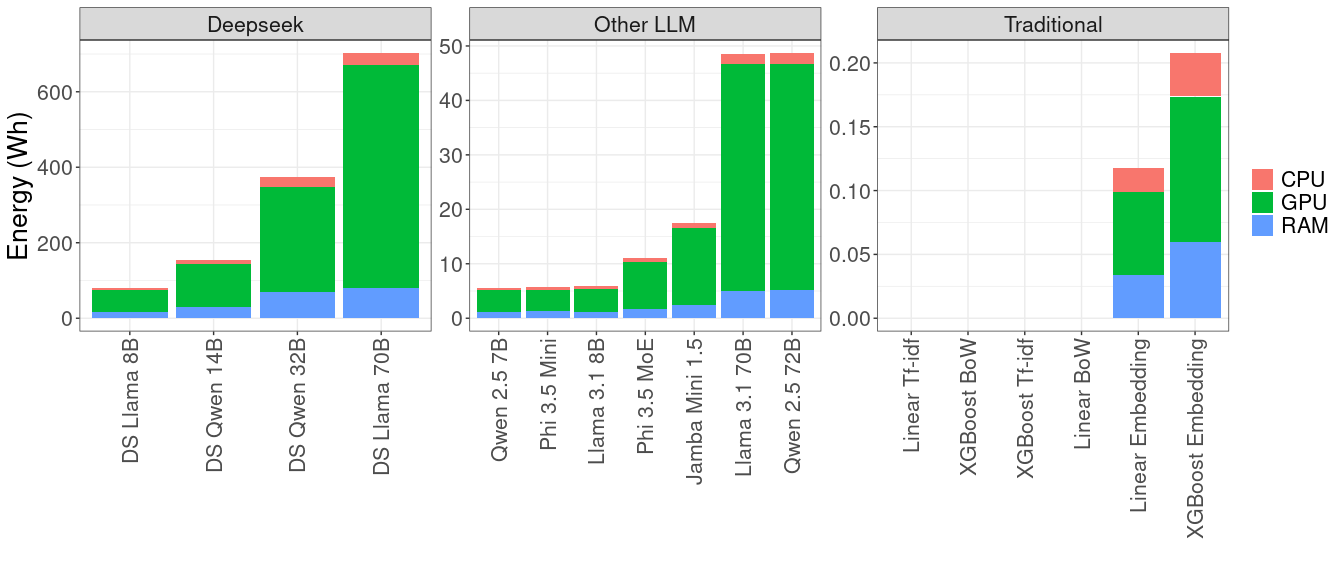}
\caption{Energy consumption of all models for the inference task on the \texttt{Capella} system (single node). All three subplots show energy in Wh (mind the different numerical ranges).}
\label{fig:inference-split}
\end{figure}

As shown in Figure \ref{fig:inference-split}, GPU consumption accounts for the largest share of total energy usage in all experiments. The only exceptions are traditional models without embeddings, which do not use the GPU during inference.

\subsubsection{Varying the Number of GPUs}

\begin{figure}[h]
\centering
\includegraphics[width=0.9\textwidth]{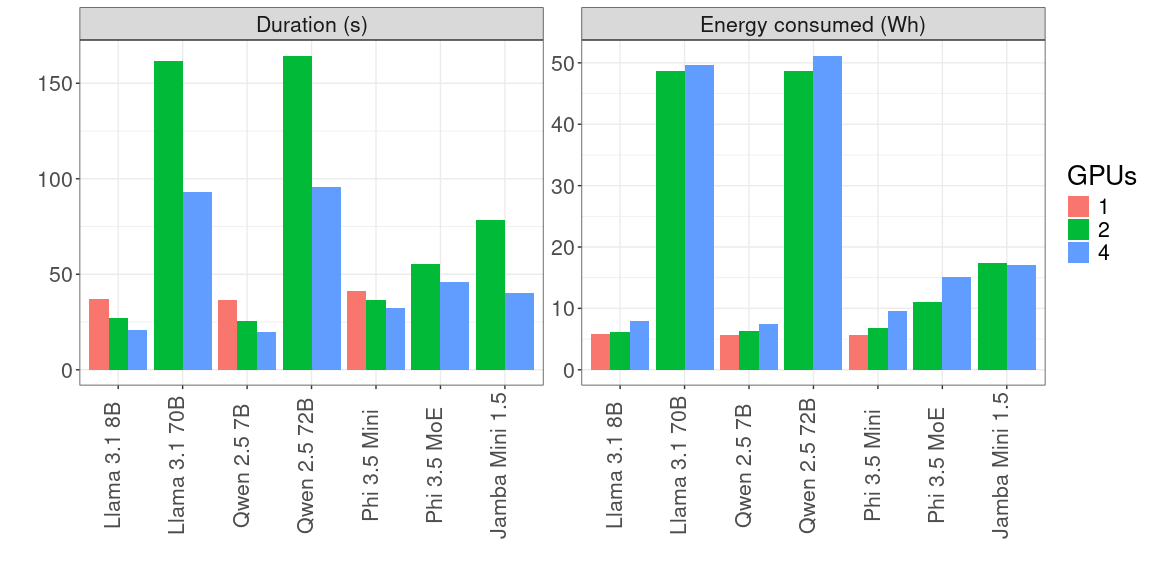}
\caption{Effects of the number of GPUs on the runtime and consumed energy (\texttt{Capella}, single node). DeepSeek models are not shown.}
\label{fig:compare-gpu-number}
\end{figure}

We examined the effect of varying the number of GPUs on energy consumption and inference duration. Most LLMs were tested on 1, 2, or 4 GPUs on a single \texttt{Capella} system node. Larger models (Qwen 72B, Phi MoE, Llama 70B, Jamba Mini, and DS Llama 70B) required either 2 or 4 GPUs. Increasing the number of GPUs consistently reduced inference duration but did not reduce energy consumption. In some cases, energy consumption increased due to the additional GPUs in operation (see Figure \ref{fig:compare-gpu-number}).

\subsubsection{Varying the Number of Nodes}

While large models can often be executed on a single computing node, certain hardware limitations or shared high-performance computing (HPC) environments may necessitate using multiple nodes. In shared systems, it is often easier to access two nodes with half the number of available GPUs than a single node with all its GPUs, due to scheduling constraints and resource allocation policies. However, deploying models across multiple nodes increases network communication overhead and significantly raises energy consumption.

\begin{figure}[h]
\centering
\includegraphics[width=0.8\textwidth]{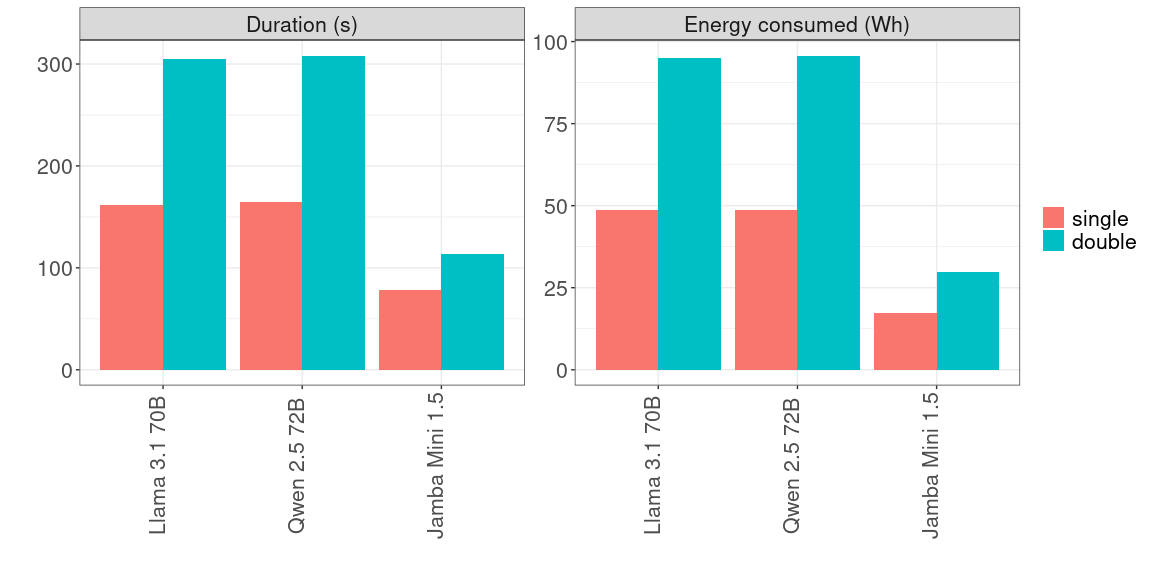}
\caption{Comparison single node vs. double node deployment (\texttt{Capella}).}
\label{fig:compare-single-double-node}
\end{figure}

We evaluated this effect for the largest models on the \texttt{Capella} system by comparing a `single-node' configuration (2 GPUs on one node) with a `double-node' configuration (1 GPU on each of two nodes). For the double-node configuration, energy consumption was summed across both nodes and averaged over 10 runs, while the reported duration reflects the average of the maximum value between the two nodes.

As shown in Figure \ref{fig:compare-single-double-node}, using two nodes increased energy consumption by a factor that depends on the model (see also Table \ref{tbl:compare-single-double-node}). This increase stems from the overhead of coordinating across nodes. Inference duration also increased by the same factor due to the sequential execution of model components and the required inter-node communication.

\subsubsection{Comparing GPU Architectures}

\begin{figure}[h]
\centering
\includegraphics[width=\textwidth]{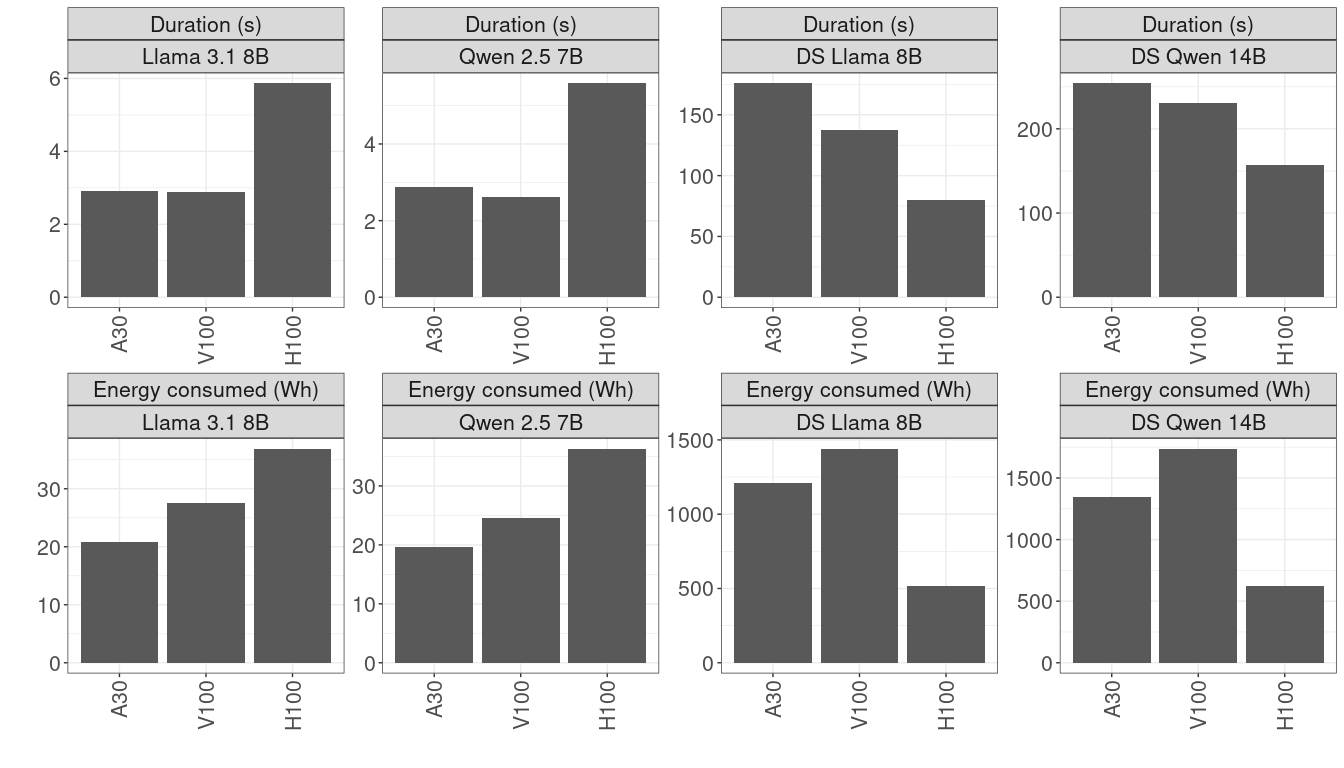}
\caption{Comparison of different GPU cards: four exemplary LLMs. Single node deployment.}
\label{fig:compare-hardware}
\end{figure}

Finally, we compared the energy efficiency of different GPU architectures (see Figure \ref{fig:compare-hardware} and Table \ref{tbl:compare-hardware}). Interestingly, the expected efficiency gains from using the more powerful H100 instead of V100 or A30 GPUs were only observed for the DeepSeek models. This discrepancy is likely to arise because DeepSeek models engage in extended reasoning by generating a larger output of words before making a classification decision. Consequently, the efficiency of H100 GPUs becomes evident only when substantial text is generated. For models generating a single token per inference, a V100 or even a A30 GPU is more efficient in inference.

\subsection{Linear relationship between duration and energy}

\begin{figure}[h]
\centering
\includegraphics[width=0.8\textwidth]{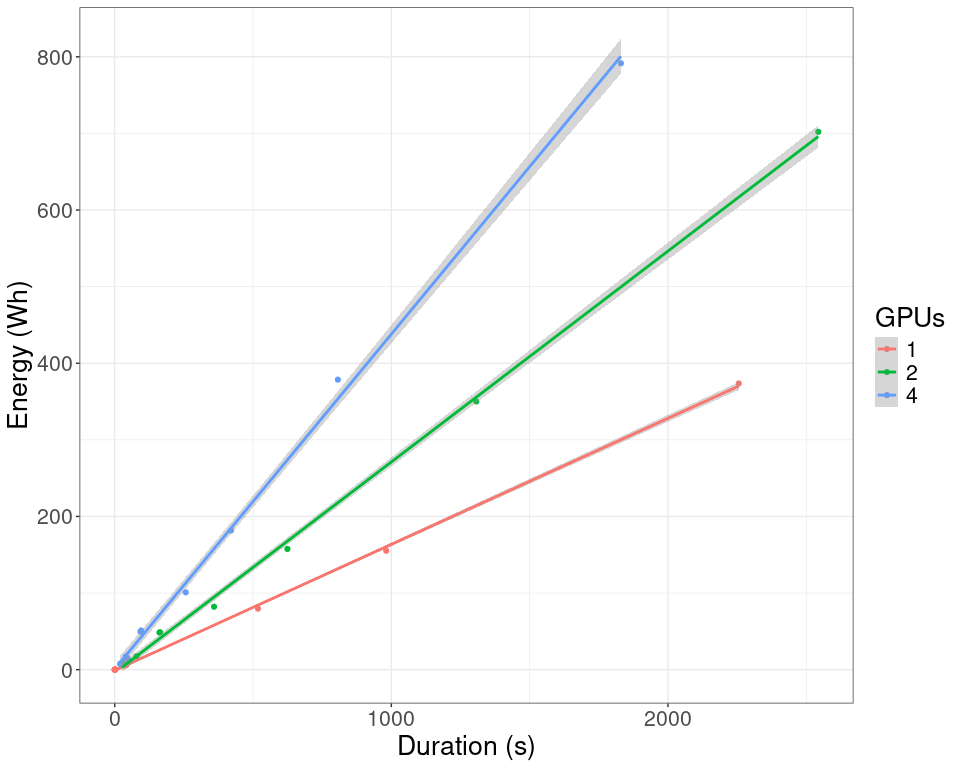}
\caption{Plotting the relationship between duration and energy consumption (single node on \texttt{Capella}). The lines are added by running a linear regression model.}
\label{fig:duration-energy}
\end{figure}

In most of the tables in appendix \ref{secApp:Tables}, we report both the duration of each inference run and its corresponding energy consumption. Since energy is the integral of power over time, these two measures exhibit a strong correlation. If the power is constant over time, this correlation should be linear. Figure \ref{fig:duration-energy} illustrates this relationship for all experiments conducted on a single node of the \texttt{Capella} cluster. When controlling for the number of GPUs used for model deployment, the relation between duration and energy is approximately linear. Therefore, the duration appears to serve as a good proxy for the energy consumed.

To further quantify the relationship between duration and energy consumption, we performed a linear regression analysis for each hardware configuration (see Table \ref{tbl:lm-duration-energy}). This analysis includes all experiments, regardless of the number of nodes used for model deployment. The consistently high R$^{2}$ values across all configurations indicate that, for a given hardware setup, duration and energy consumption are nearly interchangeable as measures of computational effort.

Moreover, when the regression coefficients are known for a specific computing system, energy consumption can be reliably estimated from the duration and the number of GPUs. Only the coefficients of duration ($a$) and of the interaction term duration:GPUs ($c$) are statistically significant. The other coefficients ($b$ and $d$) are omitted from the approximation:

\begin{equation}
\text{Energy} \approx \left( a + c \cdot \text{GPUs} \right) \cdot \text{Duration}.
\end{equation}

For instance, on the \texttt{Capella} system, the following approximation holds for any computation: 

\begin{equation}
\frac{\text{Energy}}{1\,\mathrm{Wh}} \approx \left( 0.1 + 0.09 \cdot \text{GPUs}\right) \cdot \frac{\text{Duration}}{1\,\mathrm{s}}.
\end{equation}

This relationship suggests that, under fixed hardware conditions, monitoring the duration of computations provides an efficient means of estimating energy usage with minimal additional measurement overhead.

\begin{table}[!htbp] \centering 
 
\begin{tabular}{@{\extracolsep{5pt}}lccc} 
\\[-1.8ex]\hline 
\hline \\[-1.8ex] 
 & \multicolumn{3}{c}{Dependent variable: Energy} \\ 
\cline{2-4} 
\\[-1.8ex] & \multicolumn{3}{c}{} \\ 
 & Capella & Clara & Paula \\ 
\\[-1.8ex] & (1) & (2) & (3)\\ 
\hline \\[-1.8ex] 
 Duration ($a$) & 0.097$^{***}$ & 0.061$^{***}$ & 0.079$^{***}$ \\ 
  & (0.008) & (0.002) & (0.026) \\ 
  & & & \\ 
 GPUs ($b$) & $-$0.500 & 0.048 & $-$2.195 \\ 
  & (2.297) & (0.339) & (3.472) \\ 
  & & & \\ 
 Duration:GPUs ($c$) & 0.090$^{***}$ & 0.036$^{***}$ & 0.054$^{***}$ \\ 
  & (0.004) & (0.0002) & (0.004) \\ 
  & & & \\ 
 Constant ($d$) & $-$6.205 & $-$0.826 & 3.328 \\ 
  & (5.725) & (1.368) & (17.220) \\ 
  & & & \\ 
\hline \\[-1.8ex] 
Observations & 44 & 19 & 23 \\ 
R$^{2}$ & 0.998 & 1.000 & 0.989 \\ 
Adjusted R$^{2}$ & 0.998 & 1.000 & 0.987 \\ 
\hline 
\hline \\[-1.8ex] 
\textit{Note:}  & \multicolumn{3}{r}{$^{*}$p$<$0.1; $^{**}$p$<$0.05; $^{***}$p$<$0.01} \\ 
\end{tabular} 
\caption{Linear regression of energy consumption on duration \citep[table format by][]{stargazer}. The numbers (coefficients) give the estimated effects of each predictor on the dependent variable. A positive coefficient means the variable increases the outcome, while a negative coefficient means it decreases the outcome. The standard error (in parenthesis) estimates the variability of the coefficient estimate. The p-value (given by the asterisks) indicates whether the predictor is statistically significant (different from zero).}
\label{tbl:lm-duration-energy}
\end{table} 

\section{Discussion}\label{sec12}

We would like to mention the limitations of our study, which also point to the areas of future research. First, while traditional models were trained on approximately 200 examples, the large language models (LLMs) were applied in a zero-shot setting, meaning they had no access to labeled examples. Previous research has shown that few-shot prompting - where representative examples are included in the prompt - can improve performance \citep{BrownEtAl2020}. For the present study, we kept the prompt as simple as possible (see Appendix \ref{secApp:LLMPrompt}). But in an actual application, we would add background information about the data and the categories. In general, prompt engineering, the addition of representative examples to the prompt, or even fine-tuning an LLM could yield higher accuracy rates. On the other hand, energy efficiency in LLMs can be improved through model quantization, which reduces computational demands by compressing model parameters \citep{jacobetal2017}.

Second, we do not account for the energy costs associated with training the traditional models because it is infeasible to compare them to the training costs of LLMs. The LLMs used in this study were pre-trained by external organizations and made publicly available. As a result, the energy costs of training are distributed among all users, making it difficult to estimate per-user energy consumption. Even if training energy costs for an LLM were known, the number of users remains uncertain. Additionally, hosting LLMs (e.g., on Hugging Face) and managing network traffic also contribute to energy consumption. Deploying an LLM on a dedicated server (e.g., using vLLM) requires setup time and additional energy. Beyond inference, significant time and computational resources are also required for development tasks, including data processing, testing different models and prompts, parameter tuning, and debugging - workloads that apply to both traditional models and LLMs. The measurement of additional related energy consumptions (such as network traffic or disk storage) is beyond the scope of this paper.

Third, energy consumption was measured using CodeCarbon, a tool recognized for providing reliable estimates of a machine's total energy use \citep{Bouza2023}. However, it does not allow for precise measurement of energy consumption at the level of individual processes. Moreover, power intake was recorded at 15-second intervals, meaning the accuracy of energy estimates improves with longer-running processes. As a result, energy estimates for very short-running tasks, such as inference with small traditional models, carry higher relative uncertainty, and absolute values should be interpreted with caution. Consequently, relative comparisons between very fast, CPU-bound models and slower, GPU-accelerated LLMs may be influenced by these approximations, though the overall ordering and major trends are expected to remain reliable.
Another limitation of CodeCarbon is that RAM energy consumption is approximated at 0.375W per GB of memory used. While the Running Average Power Limit (RAPL) framework can directly measure RAM power consumption, it is not supported on all CPUs\footnote{\url{https://github.com/mlco2/codecarbon/issues/717\#issuecomment-2589805160}}. Additionally, in shared computing environments such as high-performance computing (HPC) clusters, measurements may be affected by other users' activities. Especially when an LLM was deployed across multiple nodes, variations in network traffic at different times may have influenced energy measurements. A more precise assessment of energy efficiency would benefit from using dedicated computing resources with physical wattmeters and high-resolution energy measurement tools\citep[e.g.][]{IlscheEtAl2019}. While the exact power measurement on hardware level would be interesting, this is often not possible in HPC systems for administrative and organizational reasons.

In the following, we assess further limitations of the present study in more detail. More specifically, we address our focus on a single dataset in section \ref{sec:other-datasets} and the limitation to the text categorisation task in section \ref{sec:categorisation-task}. Subsequently, we contextualise our work in the broader context of planet-centered LLMs (section \ref{sec:planet-centered-llms}) and discuss policy implications (section \ref{sec:policy}).

\subsection{Analysis on other datasets}\label{sec:other-datasets}

Our analysis was conducted on a highly specialized dataset. To assess the generalizability of our findings, we replicated the experiments using four additional, widely used datasets (see table \ref{tbl-dataset-selection}). These datasets were selected from the HuggingFace platform based on popularity and had to meet two criteria: suitability for text classification and inclusion of two columns - \texttt{text} and \texttt{label}. To maintain comparability with our initial analysis, we randomly sampled 200 training examples and 200 test examples from each dataset. Using a slightly larger training set might have provided an advantage to traditional models, as the LLMs were applied in a zero-shot setting without fine-tuning. Each model experiment was repeated 10 times with different samples, ensuring that each model was tested on the same 10 sets.

Beyond expanding the dataset variety, these four benchmarks also introduce differences in label balance, linguistic complexity, and domain heterogeneity. For example, the Yelp and Rotten Tomatoes datasets exhibit strong class imbalance, while the AG News and Emotion datasets involve broader topical and semantic diversity. Including these complementary datasets allows us to evaluate our core findings under substantially more complex and imbalanced conditions, thereby analysing the generalizability of our conclusions

\begin{table}[h]
\centering
\begin{tabular}{lll}
\hline
Name & Classification Task & ID on \url{https://huggingface.co/datasets} \\ 
\hline
news & \makecell[lt]{news topics: World, Sports,\\Business, Sci/Tech} & fancyzhx/ag\_news \\ 
yelp & sentiment: 1-5 stars & Yelp/yelp\_review\_full \\ 
tomatoes & sentiment: pos, neg & cornell-movie-review-data/rotten\_tomatoes \\ 
emotion & \makecell[lt]{emotion: anger, fear, joy,\\ love, sadness, surprise} & dair-ai/emotion\\
\hline
\end{tabular}
\caption{Selection of datasets for text classification tasks.}\label{tbl-dataset-selection}
\end{table}

\begin{figure}[h]
\centering
\includegraphics[width=0.9\textwidth]{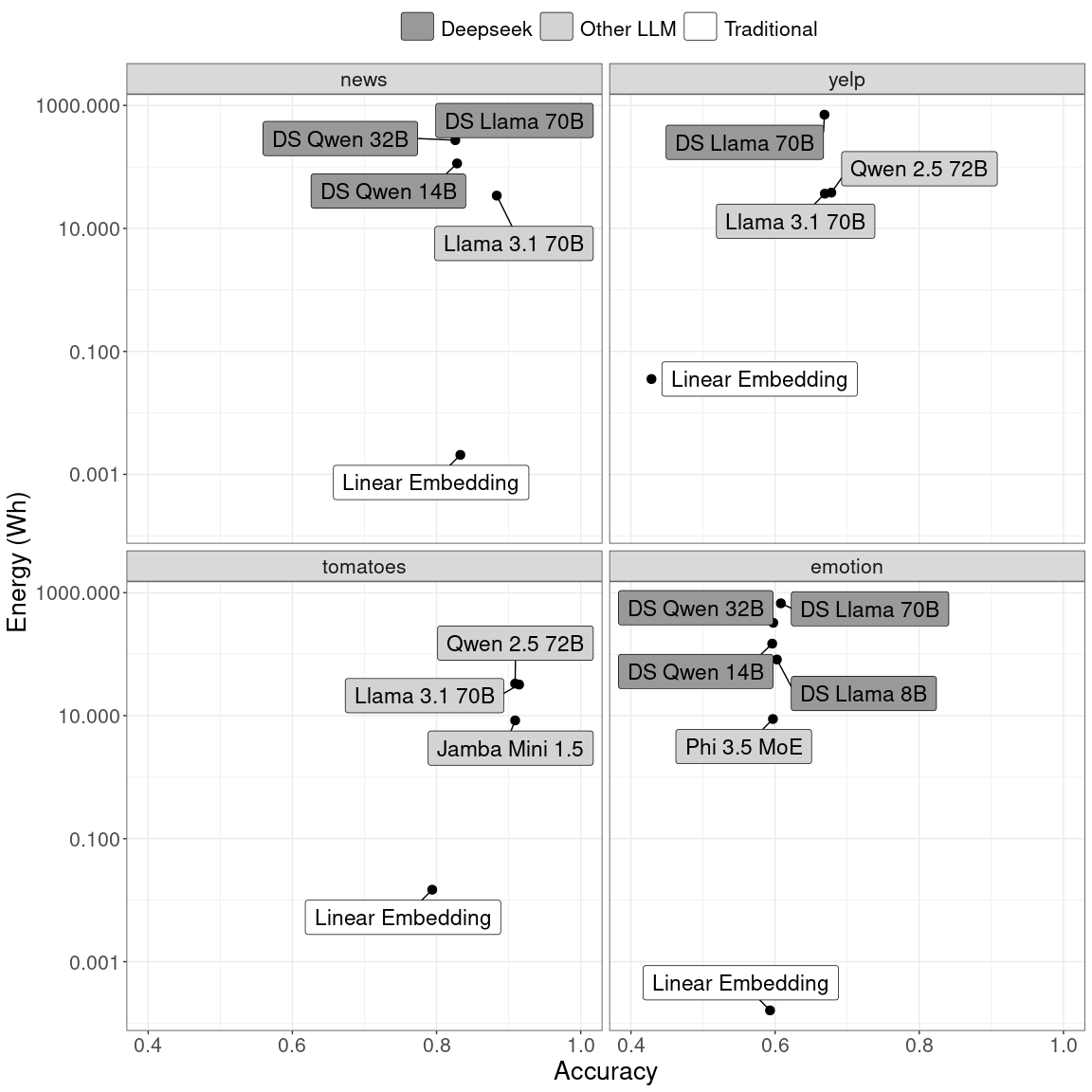}
\caption{Accuracy-energy-trade-off of the best models for the inference task on different datasets (the Linear Embedding model was added for comparison), \texttt{Capella} system, single node. See Tables \ref{tbl:inference-all-news-yelp} and \ref{tbl:inference-all-tomatoes-emotion} for results of all models.}
\label{fig:inference-other-datasets}
\end{figure}

Figure \ref{fig:inference-other-datasets} visualizes the relationship between accuracy and energy consumption for these additional text classification tasks. For clarity, we restricted the visualization to the models with the three highest accuracy scores and included the linear model with sentence embeddings for comparison (see Tables \ref{tbl:inference-all-news-yelp} and \ref{tbl:inference-all-tomatoes-emotion} for details).

Similar to our findings with the FKTG dataset, the DeepSeek models do not outperform the best non-reasoning models in most cases. The only exception is the emotion dataset, where DeepSeek Llama 70B achieves an accuracy of $0.61$, slightly surpassing the best non-reasoning model, Phi 3.5 MoE ($0.60$). However, unlike in the previous analysis, for every dataset, at least one LLM outperforms the best traditional model, demonstrating that the optimal model choice is highly task-dependent.

For the news dataset, Llama 3.1 70B achieves an accuracy $0.05$ points higher than the best linear model ($0.88$ vs. $0.83$). However, this comes at the cost of significantly higher energy consumption ($34.15$ Wh vs. $0.0021$ Wh), highlighting the need for careful trade-off considerations. In contrast, for sentiment analysis on the Yelp dataset, traditional models perform considerably worse than LLMs (accuracy differences exceeding $
0.08$), potentially justifying the energy costs of LLM deployment in this context. This shows that task characteristics - such as semantic complexity and required contextual understanding - play a crucial role in determining model suitability.

In some cases, smaller LLMs may offer acceptable performance with reduced energy consumption. For instance, Qwen 2.5 7B shows that while its accuracy is slightly lower than the version with 72B parameters ($0.60$ vs. $0.68$), it consumes only one-eighth of the energy (yelp dataset). A similar pattern is observed for sentiment analysis on the Rotten Tomatoes dataset, where traditional models again fail to match LLM performance. Among the larger models, Jamba Mini 1.5 stands out as one of the most efficient choices, offering strong accuracy while consuming significantly less energy. Notably, despite having nearly as many parameters as Llama 3.1 70B and Qwen 2.5 72B ($51.6$B vs. $70$B/$72$B), Jamba Mini 1.5 requires only a quarter of the energy for the same task.

For emotion classification, the linear model with sentence embeddings is among the top-performing models, revealing that traditional approaches remain competitive for certain classification tasks. These mixed results across datasets underscore that no single approach is universally superior; rather, accuracy-energy trade-offs must be assessed on a case-by-case basis considering task complexity, required performance levels, and available computational resources. While our findings suggest traditional models are often sufficient for straightforward classification tasks - particularly in specialized domains like FKTG - LLMs offer justifiable benefits for semantically complex tasks despite higher energy consumption. However, we note that superior LLM performance on some benchmark datasets might partly reflect data leakage during pre-training, whereas our domain-specific FKTG dataset likely falls outside typical training corpora. Importantly, across all datasets examined, test-time compute as featured by the DeepSeek models shows no consistent benefits in text classification tasks, and the linear relationship between computation runtime and energy consumption holds universally (see Table \ref{tbl:lm-duration-energy-other-datasets}).

\subsection{Transferability to other tasks}\label{sec:categorisation-task}
   
Another limitation of this study is its focus on text categorization; extending the analysis to all major NLP tasks would require multiple task-specific pipelines, evaluation metrics, and energy-measurement setups, which is beyond the scope of a single study. Our goal was instead to provide a detailed and methodologically consistent comparison within one task family, while demonstrating cross-dataset robustness. This focus allows for a straightforward measurement of a model's performance (using the accuracy metric). Recent studies suggest that similar comparisons in terms of efficiency and accuracy can be insightful in a variety of domains beyond categorization. For instance, \citet{clavie2025} demonstrate that simple encoder-based models can effectively tackle generative tasks, expanding the potential applications of smaller, less energy-hungry models. 

Moreover, a growing body of research highlights the advantages of fine-tuned small models for specialized tasks, where they often outperform larger models \citep{savvov2025}. This trend is evident in studies such as \citet{wei2024adaptedlargelanguagemodel}, where a diabetes-specific LLM - despite having significantly fewer parameters - outperforms both GPT-4 and Claude-3.5 in processing various diabetes tasks. Similarly, \citet{lu2023humanwinsllmempirical} report that their fine-tuned models achieve performance levels comparable to GPT-4 on domain-specific annotation tasks, yet with hundreds of times fewer parameters and significantly reduced computational costs. \citet{zhan2025slmmodsmalllanguagemodels} further emphasize the superior performance of fine-tuned small models over zero-shot LLMs, particularly in in-domain content moderation tasks. 

The study by \citet{LuccioniEtAl2024} provides additional insights into the balance between model size and efficiency while looking at ten different machine learning tasks including image classification and captioning, question answering, summarization, as well as image and text generation. The authors demonstrate that smaller models can achieve high performance with considerably less resource consumption. Their initiative resulted into the AI Energy Score\footnote{\url{https://huggingface.co/AIEnergyScore}}, a tool designed to assess the environmental impact of AI models on a range of tasks, and reinforces the growing importance of considering energy efficiency in model evaluation. 

\subsection{Further Requirements of Planet-Centered LLMs}\label{sec:planet-centered-llms}
While energy consumption and the associated carbon footprint remain crucial considerations for sustainable AI, truly planet-centered LLMs must meet a broader set of requirements that go beyond mere efficiency. These include other limited resources (water, rare-earth metals, landuse,\dots), transparency, accessibility, ethical considerations, and technical adaptability to ensure responsible and sustainable AI deployment \citep{falk2025carbonc}.

Transparency in AI models is essential for trust and reproducibility \citep{Raji2020}. The predictions of traditional models are generally more transparent than those of LLMs. Open-source LLMs, where both model architectures and training data are publicly available, contribute to scientific progress, allow for direct model comparisons such as this present study, and reduce dependency on proprietary technologies \citep{Wei2023}. Furthermore, the ability to inspect training data is crucial to assess potential biases and copyright compliance \citep{Bender2021}. Many proprietary models, such as GPT-4, lack such transparency, making it difficult to evaluate their fairness and ethical considerations. The EU AI Act will require providers of general-purpose AI models to publish a sufficiently detailed summary of their training data starting in August 2025, which further highlights the call for transparency.

LLMs vary significantly in size, ranging from lightweight models such as fastText \citep{Joulin2017} to massive architectures like BLOOM-176B, which require substantial GPU memory and network bandwidth \citep{luccioni2022}. These computational demands translate into high operational costs and environmental impacts. Moreover, some models require proprietary hardware, limiting their accessibility and long-term sustainability. Future AI systems should prioritize modularity and adaptability, enabling efficient integration into diverse infrastructures without excessive resource demands.

The relevance and fairness of AI-generated outputs depend on the quality and recency of training data. Stale or biased datasets can lead to misleading results and reinforce harmful stereotypes \citep{Bender2021, Gehman2020}. In particular, the presence of toxic content or hate speech in training data can result in models generating harmful or discriminatory outputs, which poses serious challenges for their deployment in sensitive contexts such as education, healthcare, or public administration. Moreover, safety concerns—such as the risk of models producing factually incorrect, manipulative, or otherwise harmful content—are especially critical in public-sector applications, where accountability and trust are paramount \citep{Weidinger2021}. Addressing these challenges requires robust bias-mitigation strategies and transparent documentation of model behavior.

To align with global sustainability and ethical AI principles, future research should emphasize the development of adaptable, transparent, and energy-efficient LLMs. By integrating principles of openness, fairness, and regulatory compliance, we can foster AI systems that not only minimize environmental impact but also promote responsible and equitable usage across sectors.

\subsection{Policy Implications}
\label{sec:policy}
The findings of this study have practical relevance for public administration IT departments that deploy high-volume, repetitive text-classification systems, such as continuous batch processing or internal document tagging. For these scenarios, small, task-specific models often achieve comparable accuracy to large LLMs while consuming orders of magnitude less energy. Consequently, energy efficiency should be considered alongside accuracy and other standard performance metrics when selecting models for deployment. This aligns with the proportionality and sustainability principles highlighted in the EU AI Act.

To support informed decision-making, agencies are encouraged to adopt a structured, internal model evaluation workflow that incorporates both accuracy and energy consumption. In line with general ML best practice, this begins with simple, traditional models as benchmarks, followed by evaluation of more complex LLMs only if they deliver a meaningful improvement in performance. During this process, energy usage should be measured and considered alongside accuracy and explainability to ensure that any gains justify the additional computational cost.

Maintaining an open-source, reusable model evaluation template can facilitate consistent and reproducible assessments across projects. We provide our model evaluation pipeline open source\footnote{\url{https://gitlab.opencode.de/uba-ki-lab/llm-testframework}} for reuse. Public institutions can also leverage community-maintained benchmarks, such as the Hugging Face AI Energy Score leaderboard\footnote{\url{https://huggingface.co/spaces/AIEnergyScore/Leaderboard}}, to stay informed about emerging models and relative energy consumption. Sharing internally developed best-practice workflows in open-source formats further enables learning across departments and encourages climate-conscious adoption of AI technologies.

In practice, this approach supports four strategic objectives: (i) prioritize explainable models that meet operational requirements, (ii) select models that balance accuracy with energy efficiency, (iii) adopt more complex LLMs only when justified by measurable benefits, and (iv) maintain adaptability to evolving AI tools through periodic review and benchmarking. Collectively, these measures enable cost-effective, climate-responsible digitalization without imposing additional bureaucratic or regulatory burden.

\bmhead{Acknowledgements}
We gratefully acknowledge the support provided by the Federal Ministry for the Environment, Nature Conservation and Nuclear Safety (BMUV). Additionally, we thank colleagues from Z 2.3 and  the entire AI-Lab team for their support and inspiration. We appreciate the constructive feedback from various anonymous referees that helped to improve this manuscript.

This work was supported by high-performance computer time and resources from the Center for Information Services and High Performance Computing (ZIH) of TUD Dresden University of Technology and the systems for scientific computing of Leipzig University. We thank the Center for Scalable Data Analytics and Artificial Intelligence (ScaDS.AI Dresden/Leipzig) for their support in the acquisition process.

The tool ChatGPT (OpenAI) was used to revise the text of the paper.

\bmhead{Author contribution statements}
T.H. conceived the study, initiated the project, led the research effort, and contributed to the literature review and manuscript writing. J.Z. designed and implemented the experiments, developed the codebase, conducted data analysis, and contributed to drafting the manuscript.

\bmhead{Competing interests} There are no competing interests.

\bmhead{Funding declaration} This research did not receive any specific grant from funding agencies in the public, commercial, or non-profit sectors. The work was carried out as part of the authors' regular employment duties.

\bmhead{Availability of data and code} The source code and all underlying data is public: \url{https://gitlab.opencode.de/uba-ki-lab/llm-testframework}.

%
%
%
%
%
%
%

\newpage

\begin{appendices}





\section{LLM prompt}\label{secApp:LLMPrompt}

For the zero-shot classification, we prompted the LLM with the following instruction (originally in German):

\lstset{texcl=true,basicstyle=\small\sf,commentstyle=\small\rm,mathescape=true,escapeinside={(*}{*)}}
\begin{lstlisting}
Classify the text as one of the following categories:
- <category 1>
- <category 2>
- ...
\end{lstlisting}

The categories were a fixed set of 14 options that occurred in the training as well as the test dataset: `geoWK', `Tongestein', `Aktive Störungszonen', `Öffentlichkeitsbeteiligung', `Kristallingestein', `FEP/Szenarien/Entwicklungen des Endlagersystems', `Anwendung geoWK', `Mindestanforderungen', `Steinsalz in steiler Lagerung', `Datenverfügbarkeit', `Modellierung', `Referenzdatensätze', `Bereitstellung der Daten', `Ausschlusskriterien'.

Since we deployed the \texttt{dspy} framework (\url{https://dspy.ai/}) to query the LLMs, the final prompt was automatically extended to the following:

\begin{lstlisting}
- role: system
  content: |-
    Your input fields are:
    1. `text` (str)

    Your output fields are:
    1. `category` (str)

    All interactions will be structured in the following way, 
    with the appropriate values filled in.

    [[ ## text ## ]]
    {text}

    [[ ## category ## ]]
    {category}

    [[ ## completed ## ]]

    In adhering to this structure, your objective is: 
            Classify the text as one of the following categories:
            - <category 1>
            - <category 2>
            - ...
- role: user
  content: |-
    [[ ## text ## ]]
    <text>

    Respond with the corresponding output fields, starting with 
    the field `[[ ## category ## ]]`, and then ending with the 
    marker for `[[ ## completed ## ]]`.
\end{lstlisting}

\section{Tables}\label{secApp:Tables}

\begin{table}[h]
\centering
\begin{tabular}{lrrrrr}
  \hline
Model & GPUs & Energy (Wh) & Accuracy & Duration (s) & Average Power (W) \\ 
  \hline
Linear BoW & 1 & $<$0.01 & 0.43 & 0.01 & 139.96 \\ 
  Linear Tf-idf & 1 & $<$0.01 & 0.41 & 0.01 & 43.72 \\ 
  Linear Embedding & 1 & 0.12 & 0.57 & 1.64 & 259.41 \\ 
  XGBoost BoW & 1 & $<$0.01 & 0.35 & 0.01 & 63.32 \\ 
  XGBoost Tf-idf & 1 & $<$0.01 & 0.47 & 0.01 & 67.77 \\ 
  XGBoost Embedding & 1 & 0.21 & 0.47 & 2.87 & 259.94 \\ 
  Llama 3.1 8B & 1 & 5.86 & 0.35 & 36.88 & 572.49 \\ 
  Llama 3.1 70B & 2 & 48.60 & 0.48 & 161.59 & 1082.82 \\ 
  Qwen 2.5 7B & 1 & 5.58 & 0.45 & 36.28 & 553.84 \\ 
  Qwen 2.5 72B & 2 & 48.66 & 0.51 & 164.44 & 1065.31 \\ 
  Phi 3.5 Mini & 1 & 5.74 & 0.30 & 41.45 & 498.46 \\ 
  Phi 3.5 MoE & 2 & 11.00 & 0.40 & 55.51 & 713.34 \\ 
  Jamba Mini 1.5 & 2 & 17.42 & 0.34 & 78.61 & 797.94 \\ 
  DS Llama 8B & 1 & 79.64 & 0.37 & 517.83 & 553.67 \\ 
  DS Llama 70B & 2 & 702.06 & 0.46 & 2543.47 & 993.68 \\ 
  DS Qwen 14B & 1 & 155.20 & 0.39 & 981.35 & 569.33 \\ 
  DS Qwen 32B & 1 & 373.56 & 0.45 & 2255.99 & 596.11 \\ 
   \hline
\end{tabular}
\caption{Measurements of all models for the inference task on the FKTG dataset, \texttt{Capella} system, single node, shown are averages over 10 runs}
\label{tbl:inference-all}
\end{table}

\begin{table}[h]
\centering
\begin{tabular}{lrrrrrr}
\hline
Model & \multicolumn{3}{c}{Duration (s)} & \multicolumn{3}{c}{Energy consumed (Wh)} \\ 
 & single & double & ratio & single & double & ratio \\ 
\hline
Llama 3.1 70B & 161.59 & 304.77 & 1.89 & 48.60 & 94.88 & 1.95 \\ 
Qwen 2.5 72B & 164.44 & 308.16 & 1.87 & 48.66 & 95.70 & 1.97 \\ 
Jamba Mini 1.5 & 78.61 & 113.88 & 1.45 & 17.42 & 29.81 & 1.71 \\ 
DS Llama 70B & 2543.47 & 6792.54 & 2.67 & 702.06 & 1899.86 & 2.71 \\ 
\hline
\end{tabular}
\caption{Comparison single vs. double node deployment, \texttt{Capella} system}
\label{tbl:compare-single-double-node}
\end{table}

\begin{table}[h]
\centering
\begin{tabular}{lrrrrrr}
\hline
Model & \multicolumn{3}{c}{Duration (s)} & \multicolumn{3}{c}{Energy consumed (Wh)} \\ 
 & A30 & V100 & H100 & A30 & V100 & H100 \\ 
\hline
Llama 3.1 8B & 20.78 & 27.52 & 36.88 & 2.91 & 2.88 & 5.86 \\ 
Qwen 2.5 7B & 19.58 & 24.64 & 36.28 & 2.87 & 2.63 & 5.58 \\ 
Phi 3.5 Mini & 19.18 & 25.02 & 41.45 & 2.65 & 2.50 & 5.74 \\ 
Phi 3.5 MoE & 77.60 & 32.53 & 45.93 & 17.77 & 6.04 & 15.04 \\ 
DS Llama 8B & 1210.90 & 1439.58 & 517.83 & 175.83 & 137.90 & 79.64 \\ 
DS Qwen 14B & 1348.09 & 1736.21 & 624.38 & 254.01 & 230.72 & 157.58 \\ 
DS Qwen 32B & 1688.23 & 2192.53 & 806.68 & 444.67 & 457.60 & 378.58 \\ 
   \hline
\end{tabular}
\caption{Comparison of different GPU cards, single node deployment.}
\label{tbl:compare-hardware}
\end{table}

\begin{table}[ht]
\centering
\begin{tabular}{lrrrr}
  \hline
Dataset & \multicolumn{2}{c}{news} & \multicolumn{2}{c}{yelp}\\ 
Model & Energy (Wh) & Accuracy & Energy (Wh) & Accuracy \\ 
  \hline
Linear BoW & $<$0.01 & 0.65 & $<$0.01 & 0.36 \\ 
  Linear Tf-idf & $<$0.01 & 0.65 & $<$0.01 & 0.34 \\ 
  Linear Embedding & $<$0.01 & 0.83 & 0.04 & 0.43 \\ 
  XGBoost BoW & $<$0.01 & 0.48 & $<$0.01 & 0.31 \\ 
  XGBoost Tf-idf & $<$0.01 & 0.52 & $<$0.01 & 0.29 \\ 
  XGBoost Embedding & 0.03 & 0.74 & 0.01 & 0.40 \\ 
  Llama 3.1 8B & 4.31 & 0.71 & 4.73 & 0.58 \\ 
  Llama 3.1 70B & 34.15 & 0.88 & 36.71 & 0.67 \\ 
  Qwen 2.5 7B & 4.21 & 0.01 & 4.52 & 0.60 \\ 
  Qwen 2.5 72B & 33.75 & 0.79 & 38.20 & 0.68 \\ 
  Phi 3.5 Mini & 3.30 & 0.53 & 15.55 & 0.58 \\ 
  Phi 3.5 MoE & 8.53 & 0.78 & 8.32 & 0.58 \\ 
  Jamba Mini 1.5 & 9.34 & 0.78 & 11.45 & 0.56 \\ 
  DS Llama 8B & 60.58 & 0.82 & 97.18 & 0.62 \\ 
  DS Llama 70B & 483.73 & 0.83 & 707.03 & 0.67 \\ 
  DS Qwen 14B & 113.81 & 0.83 & 177.41 & 0.63 \\ 
  DS Qwen 32B & 271.92 & 0.83 & 358.62 & 0.63 \\ 
   \hline
\end{tabular}
\caption{Measurements of all models for the inference task on the news and yelp datasets, \texttt{Capella} system, single node, shown are averages over 10 runs}
\label{tbl:inference-all-news-yelp}
\end{table}

\begin{table}[ht]
\centering
\begin{tabular}{lrrrr}
  \hline
Dataset & \multicolumn{2}{c}{tomatoes} & \multicolumn{2}{c}{emotion}\\ 
Model & Energy (Wh) & Accuracy & Energy (Wh) & Accuracy \\ 
  \hline
Linear BoW & $<$0.01 & 0.59 & $<$0.01 & 0.36 \\ 
  Linear Tf-idf & $<$0.01 & 0.59 & $<$0.01 & 0.40 \\ 
  Linear Embedding & 0.01 & 0.79 & $<$0.01 & 0.59 \\ 
  XGBoost BoW & $<$0.01 & 0.54 & $<$0.01 & 0.30 \\ 
  XGBoost Tf-idf & $<$0.01 & 0.55 & $<$0.01 & 0.33 \\ 
  XGBoost Embedding & $<$0.01 & 0.76 & $<$0.01 & 0.53 \\ 
  Llama 3.1 8B & 4.12 & 0.87 & 4.46 & 0.56 \\ 
  Llama 3.1 70B & 32.07 & 0.91 & 34.12 & 0.58 \\ 
  Qwen 2.5 7B & 4.04 & 0.73 & 4.17 & 0.37 \\ 
  Qwen 2.5 72B & 33.25 & 0.91 & 34.81 & 0.58 \\ 
  Phi 3.5 Mini & 7.20 & 0.87 & 5.13 & 0.53 \\ 
  Phi 3.5 MoE & 7.72 & 0.89 & 8.82 & 0.60 \\ 
  Jamba Mini 1.5 & 8.37 & 0.91 & 10.22 & 0.56 \\ 
  DS Llama 8B & 72.15 & 0.83 & 81.82 & 0.60 \\ 
  DS Llama 70B & 510.86 & 0.90 & 670.40 & 0.61 \\ 
  DS Qwen 14B & 134.02 & 0.89 & 148.20 & 0.60 \\ 
  DS Qwen 32B & 246.48 & 0.89 & 323.48 & 0.60 \\ 
   \hline
\end{tabular}
\caption{Measurements of all models for the inference task on the tomatoes and emotion datasets, \texttt{Capella} system, single node, shown are averages over 10 runs}
\label{tbl:inference-all-tomatoes-emotion}
\end{table}

\begin{table}[!htbp] \centering 
\begin{tabular}{@{\extracolsep{5pt}}lcccc} 
\\[-1.8ex]\hline 
\hline \\[-1.8ex] 
 & \multicolumn{4}{c}{Dependent variable: Energy} \\ 
\cline{2-5} 
\\[-1.8ex] & \multicolumn{4}{c}{} \\ 
 & tomatoes & emotion & news & yelp \\ 
\\[-1.8ex] & (1) & (2) & (3) & (4)\\ 
\hline \\[-1.8ex] 
 Duration & 0.040$^{***}$ & 0.043$^{***}$ & 0.052$^{***}$ & 0.045$^{***}$ \\ 
  & (0.002) & (0.002) & (0.003) & (0.003) \\ 
  & & & & \\ 
 GPUs & $-$0.079 & $-$0.052 & 0.536 & 0.810 \\ 
  & (0.950) & (1.011) & (1.470) & (1.545) \\ 
  & & & & \\ 
 Duration:GPUs & 0.122$^{***}$ & 0.120$^{***}$ & 0.115$^{***}$ & 0.120$^{***}$ \\ 
  & (0.001) & (0.001) & (0.002) & (0.002) \\ 
  & & & & \\ 
 Constant & $-$0.397 & $-$0.464 & $-$1.300 & $-$1.773 \\ 
  & (1.290) & (1.372) & (1.985) & (2.103) \\ 
  & & & & \\ 
\hline \\[-1.8ex] 
Observations & 17 & 17 & 17 & 17 \\ 
R$^{2}$ & 1.000 & 1.000 & 1.000 & 1.000 \\ 
Adjusted R$^{2}$ & 1.000 & 1.000 & 1.000 & 1.000 \\ 
\hline 
\hline \\[-1.8ex] 
\textit{Note:}  & \multicolumn{4}{r}{$^{*}$p$<$0.1; $^{**}$p$<$0.05; $^{***}$p$<$0.01} \\ 
\end{tabular} 
\caption{Linear regression of energy consumption on duration for the datasets of section \ref{sec:other-datasets} \citep[table format by][]{stargazer}.}
\label{tbl:lm-duration-energy-other-datasets}
\end{table}

\end{appendices}

\ \\
\newpage
\ \\
\newpage


\bibliographystyle{apalike}  
\bibliography{sn-bibliography}

\end{document}